\documentclass{article}

\usepackage[final]{neurips_data_2021}

\usepackage[utf8]{inputenc} 
\usepackage[T1]{fontenc}    
\usepackage{hyperref}       
\usepackage{url}            
\usepackage{booktabs}       
\usepackage{amsfonts}       
\usepackage{nicefrac}       
\usepackage{microtype}      
\usepackage{xcolor}         
\usepackage{graphicx}
\usepackage{graphics}
\usepackage{amsmath}
\usepackage{multirow}
\usepackage{enumitem}

\newcommand{\nop}[1]{}

\title{A Dataset for Answering Time-Sensitive Questions}

\author{%
  Wenhu Chen, Xinyi Wang, William Yang Wang \\
  Department of Computer Science\\
  University of California, Santa Barbara\\
  \texttt{wenhuchen@ucsb.edu}, \texttt{xinyi\_wang@ucsb.edu}, \texttt{william@cs.ucsb.edu}
}

\begin{document}

\maketitle

\begin{abstract}
Time is an important dimension in our physical world. Lots of facts can evolve with respect to time. For example, the U.S. President might change every four years. Therefore, it is important to consider the time dimension and empower the existing QA models to reason over time. However, the existing QA datasets contain rather few time-sensitive questions, hence not suitable for diagnosing or benchmarking the model's temporal reasoning capability. In order to promote research in this direction, we propose to construct a time-sensitive QA dataset. The dataset is constructed by 1) mining time-evolving facts from WikiData and aligning them to their corresponding Wikipedia page, 2) employing crowd workers to verify and calibrate these noisy facts, 3) generating question-answer pairs based on the annotated time-sensitive facts. Our dataset poses challenges in the aspect of both temporal understanding and temporal reasoning. We evaluate different SoTA long-document QA systems like BigBird and FiD on our dataset. The best-performing model FiD can only achieve 46\% accuracy, still far behind the human performance of 87\%. We demonstrate that these models are still lacking the ability to perform consistent temporal reasoning. Therefore, we believe that our dataset could serve as a benchmark to develop NLP models more sensitive to temporal shifts. The dataset and code are released in~\url{https://github.com/wenhuchen/Time-Sensitive-QA}.
\end{abstract}

\section{Introduction}
As time evolves, many facts will evolve along with it, such as `the U.S. President', `Home Team of Lebron James', etc. Understanding the scope and interval of knowledge is an essential task studied by previous literature~\citep{allen1983maintaining}. The fact evolution is commonly reflected in our daily text corpora like Wikipedia or Daily News. For example, the Wikipedia of `Lebron James'\footnote{\url{https://en.wikipedia.org/wiki/LeBron_James}} covers the whole evolution of his home team. The temporal transition of these facts is normally scattered across the long document in very diverse expressions, either represented explicitly or implicitly. Such characteristics pose great challenges to the existing NLP models. For example, a user might pose a question like `Which team did Lebron James play for in 2007?’. The only valid evidence in Wikipedia is `Cleveland Cavaliers (2003–2010)'. To answer this question, the model is required to perform temporal reasoning. We simulate these time-sensitive trivia questions and present them to the current state-of-art QA models~\citep{zaheer2020big,izacard2020leveraging} trained on large-scale datasets. These models can only achieve a compromised 27\% accuracy, much lower than their performance on Natural Questions~\citep{kwiatkowski2019natural} with 64\% accuracy and SQuAD~\citep{rajpurkar2016squad,rajpurkar2018know} with 90\% accuracy. This gap indicates the difficulties to handle temporal questions.

In this paper, we are specifically interested in handling these temporal questions. We formally define these questions as \textbf{time-sensitive questions} based on the following criterion: a) the question contains a [time specifier] like `in 2007' or `before 2010', b) modifying the [time specifier] will lead to answer change. c) such questions require temporal reasoning. We found that these time-sensitive questions despite their ubiquity are under-studied in the existing QA datasets. For example, our human study reveals that Natural Questions~\citep{kwiatkowski2019natural} only contains less than 5\% of questions with [time specifier]. In SQuAD~\citep{rajpurkar2016squad,rajpurkar2018know}, though there is a larger portion of questions with [time specifier], these questions usually copy the original time-specifying phrases from the passage without requiring any temporal reasoning, thus not meeting condition c). In TriviaQA/WebQuestions/WebComplexQuestions~\citep{berant2013semantic,bao2016constraint,joshi2017triviaqa}, our human study reveals that there are more questions involving time specifiers, however, most of these specifiers in these questions are not modifiable. An example is `What is the title of the last Harry Potter novel in 2007', where `last Harry Potter' already implies `year=2007', therefore, the [time specifier] is redundant and not modifiable. Thus, these questions cannot be considered time-sensitive due to condition b). 

\begin{figure}[!t]
    \centering
	\includegraphics[width=1.0\linewidth]{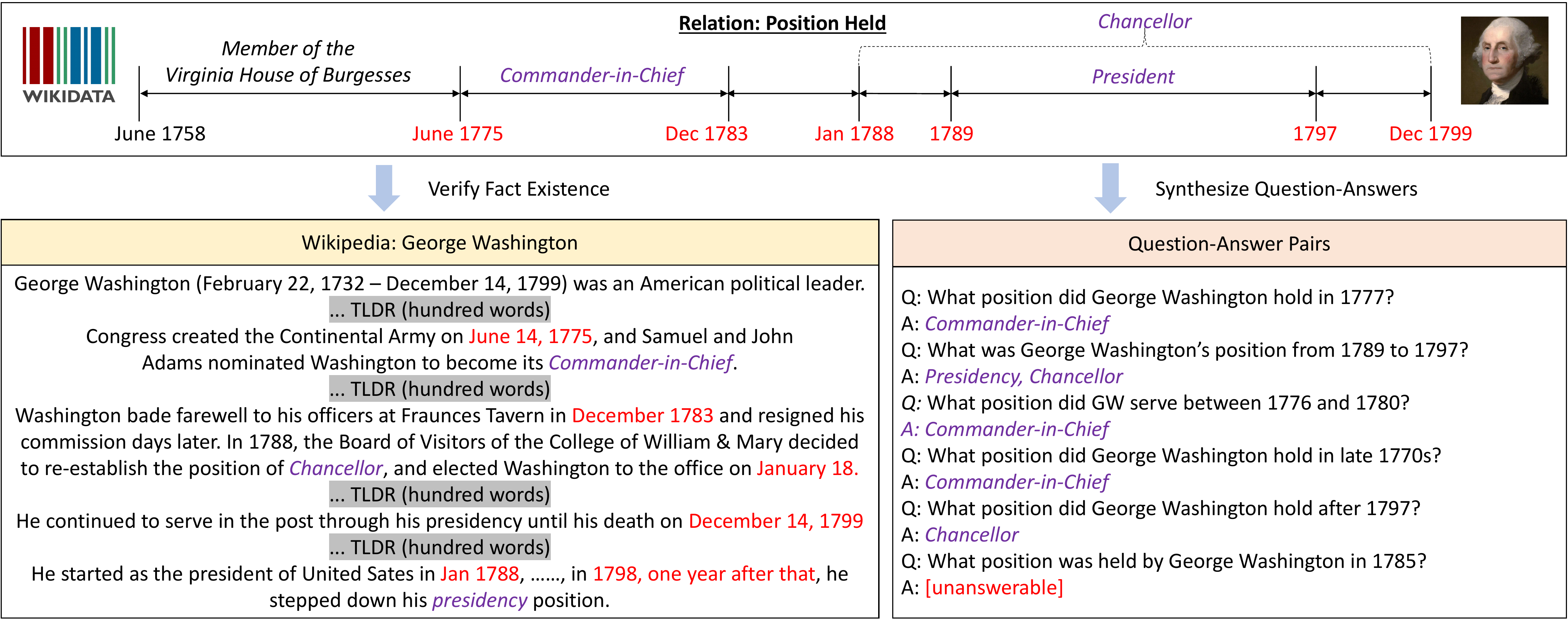}
    \caption{Time Sensitive Questions from TimeQA. } 
    \label{fig:intro}
\end{figure}

The closest to ours is Tempquestions~\citep{jia2018tempquestions,jia2018tequila}, which investigates the temporal questions with time specifiers. However, their questions are extracted from the above-mentioned datasets, which fails to meet the condition b). Furthermore, Tempquestions studies KG-based QA instead of Text-based QA, which differentiates from our goal of understanding temporal transition in natural text. Therefore, we propose to construct our own dataset called Time-Sensitive Question Answering (TimeQA). We first identify time-evolving facts from WikiData~\citep{vrandevcic2014wikidata}, and then employ human workers to annotate the boundaries of these facts by aligning with Wikipedia passages. We synthesize diverse question-answer pairs based on the annotated time-evolving facts using diverse templates. Finally, we create two datasets (easy and hard) with two levels of difficulty, both containing 20K question-answer pairs regarding 5.5K time-evolving fact and 70 relations. The hard version is more challenging as it requires more temporal reasoning than the easy version. 
The example in~\autoref{fig:intro} showcases some examples in our TimeQA dataset. The challenges posed by our dataset is in two folds:
\begin{itemize}[leftmargin=*]
    \item \textbf{Temporal Understanding}: understand the time scope (start and end time) of facts in the long text. However, the time information can be expressed implicitly in the text, which requires temporal commonsense to understand, for example, `during the second world war' implies `from 1939 to 1945', `one year after 1934' refers to `year 1944', etc.
    \item \textbf{Temporal Reasoning}: reason over the temporal information in the text conditioned on the query. More formally, the model needs to understand the temporal relationship (`within', `between', `before', `after', etc.) between the time presented in the query and document. 
\end{itemize}

We evaluate different state-of-the-art QA models' performance on both easy and hard versions, the performance drops from 60\% to 45\% on the hard version, which indicates the model suffers from its incompetence to perform temporal reasoning. When comparing with human performance of 87\% on TimeQA-hard, the existing neural models are still significantly lagging behind. Therefore, we believe TimeQA could serve as a valuable benchmark in studying this problem.

\section{Dataset and Problem Definition}
\begin{figure*}[!t]
    \centering
	\includegraphics[width=1.\linewidth]{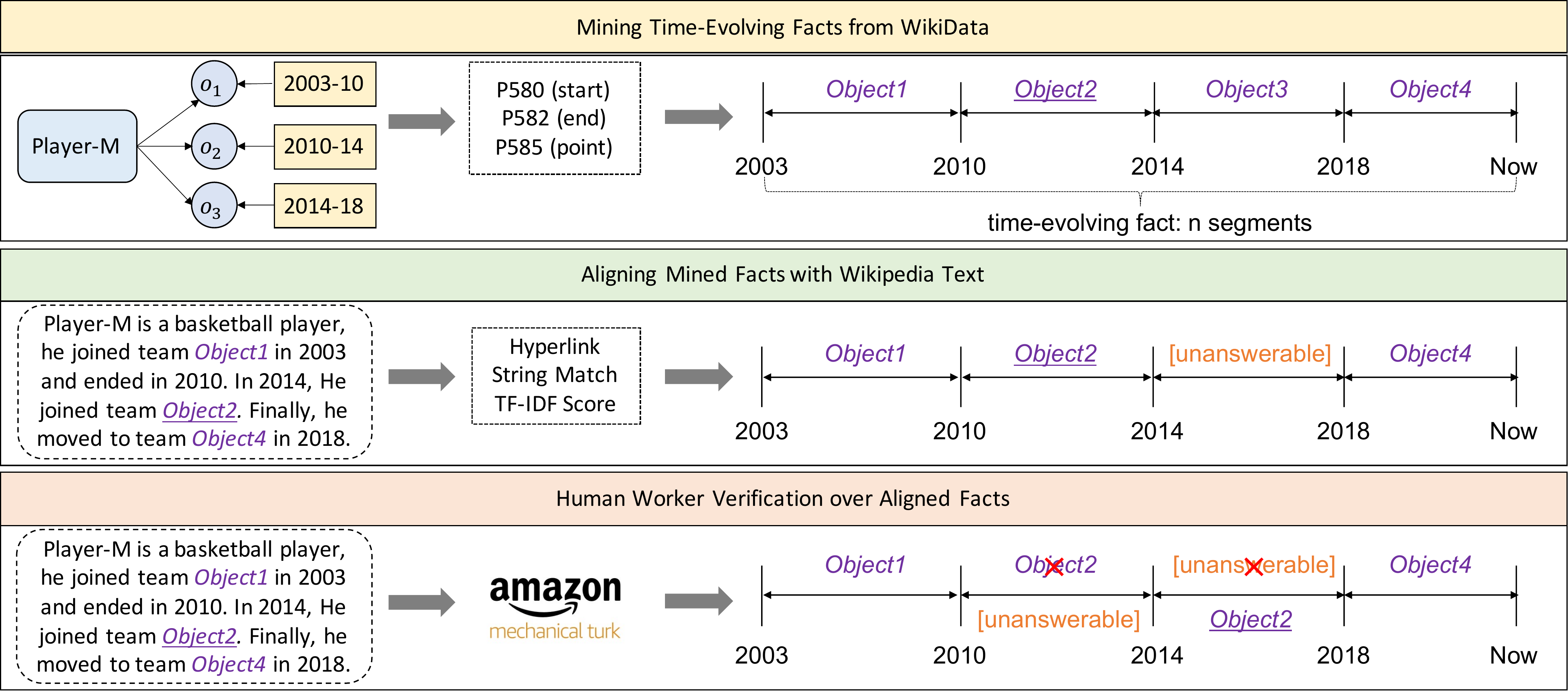}
    \caption{Fact Annotation Step: fact mining, aligning and verification.} 
    \label{fig:procedure}
\end{figure*}
Here we demonstrate our dataset construction pipeline, our dataset construction takes two steps: 1) fact annotation, 2) question-answer synthesizing.

\subsection{Fact Annotation}
The fact annotation takes three steps as depicted in~\autoref{fig:procedure}, which is comprised of three sub-steps, mining facts, aligning facts, and verify facts. 
\paragraph{Mining Time-evolving Facts from WikiData}
We first identify which facts are evolving over time and through the existing annotations from WikiData~\citep{vrandevcic2014wikidata}. Therefore, we resort to WikiData to mine these time-evolving facts, please refer to the `Lebron James' example in \url{https://www.wikidata.org/wiki/Q36159}. As indicated in the first-row of~\autoref{fig:procedure}, we traverse the WikiData pages and select the facts with time quantifier P580 (start from), P582 (end in), and P585 (point in time) to mine interesting knowledge triples with their temporal quantifiers. We discard the triples with numeric objects like ``(Denver, population of county, 13876)" because these numerical facts are unlikely to appear in Wikipedia text. We succeed to mine roughly 150K time-evolving facts in the form of ($subject$, $relation$, \{$t_1 \rightarrow t_2$: $object_1$, $t_2 \rightarrow t_3$: $object_2$, ... $t_{k} \rightarrow t_{k+1}$: $object_k$\}), where $t_k$ denotes the time boundary of $k$-th fact segment. 

\paragraph{Aligning Facts with Text}
After mining these time-evolving facts, we need to trace them back to their Wikipedia text. We decide whether the fact segment ($subject$, $relation$, $t_{k} \rightarrow t_{k+1}$: $object_k$) is mentioned in the Wikipedia page using the following rules: 1) $object_k$ is hyperlinked in the Wikipedia page, 2) $object_k$'s name has exact string match, 3) the TF-IDF score between the $object_k$'s name and Wikipedia text is above 40\%. 4) Otherwise, the segment will be deemed ``[unanswerable]". The process is depicted in the second-row of~\autoref{fig:procedure}. After this automatic tracing procedure, we discard the time-evolving facts having over than 50\% ``[unanswerable]" segments. On the other hand, we found that some particular relations like `play for' are dominating the dataset. Hence, we further propose to down-sample these over-represented relations and finally identify 5.5K well-balanced facts as our candidate knowledge triples for next-step human verification. 

\paragraph{Human Verification}
The previous step generates relatively noisy (text, time-evolving fact) pairs. There mainly exist the following sources of errors: 1) the WikiData annotation is erroneous, 2) the object is mentioned in the text, but its time boundary is not mentioned, 3) the object takes a different surface form, therefore the detected `[unanswerable]' is indeed answerable. Therefore, we propose to add human verification to clean these noisy data pairs. We provide the 5K (text, time-evolving fact) pairs as HITs to high-quality Amazon Mechanical Turkers\footnote{We select the workers from English-speaking countries, with acceptance rate over 98\% and over 1000 annotation HITs being approved.}. The workers can take the following actions: a) correcting the erroneous object, b) changing object to [unanswerable], c) changing [unanswerable] to an object in the text. The process is depicted in the third-row of~\autoref{fig:procedure}. Each HIT is paid with 1.0 dollars with an average finish time of 5 minutes. The average hourly pay is 12.0 dollars, which exceeds the income requirements proposed in human subject research protocols\footnote{\url{https://en.wikipedia.org/wiki/Minimum_wage_in_the_United_States}}. The annotation interface is demonstrated in the Appendix.

\paragraph{Annotation Criteria}
We make following guideline during annotation to help crowd-workers deal with ambivalent cases. We define extracted $t1 \rightarrow t2$ to be the correct time scope for the given fact ($subject$, $relation$, $object$) under the following conditions: a) $t_1$, $t_2$ are all explicitly mentioned in the passage, b) $t_1'$ and $t_2'$ is mentioned in the passage and $t_1' \leq t_1 < t_2 \leq t_2'$. c) $t_1',t_2'$ is more coarse grained than $t_1,t_2$ (passage mentions 2018, the question mentions 2018 June), d) $t_1', t_2'$ is mentioned in the passage, by combining with a reasoning function $f$, we are able to derive $t_1,t_2=f(t_1',t_2')$. For example, `X happens in 1987 ($t_1$), after two years, Y ...' entails `Y happening in 1989 ($t_1$)', similarly, `X went to university in 1987 ($t_2'$)' entails `X graduated in 1991 ($t_2$)'. If none of the above conditions are met, we will define the time scope to be `unknown', later on, these facts will be used to synthesize unanswerable questions.

\paragraph{Quality Control}
In order to harvest a high-quality dataset, we perform very detailed quality control in the collection procedure. In the interface, we will highlight the mentioned objects and time with special fonts to help the annotators identify them. We batch the HITs by their worker id to accelerate our quality assessment procedure. We sample 2 HITs from each batch and send them to our high-quality verifier to evaluate their correctness. If the sampled HITs pass our quality assessment, we will accept the other HITs within the batch. Otherwise, we will reject the whole batch. The overall acceptance rate is maintained between 85-90\%.

During the verification step, roughly 41.6\% of the segments are revised. The 5.5K worker-annotated examples are further filtered to obtain 5060 `golden' (text, time-evolving fact) pairs as the final release, with each fact having an average of 4 segments (time span). Out of these facts, 12\% of the segments are `[unanswerable]', 7\% have multiple objects, and 80\% have exactly one object. The harvested facts involve over 60 different relations like `play for', `employee of', etc. 


\begin{figure}[!t]
    \centering
	\includegraphics[width=1.0\linewidth]{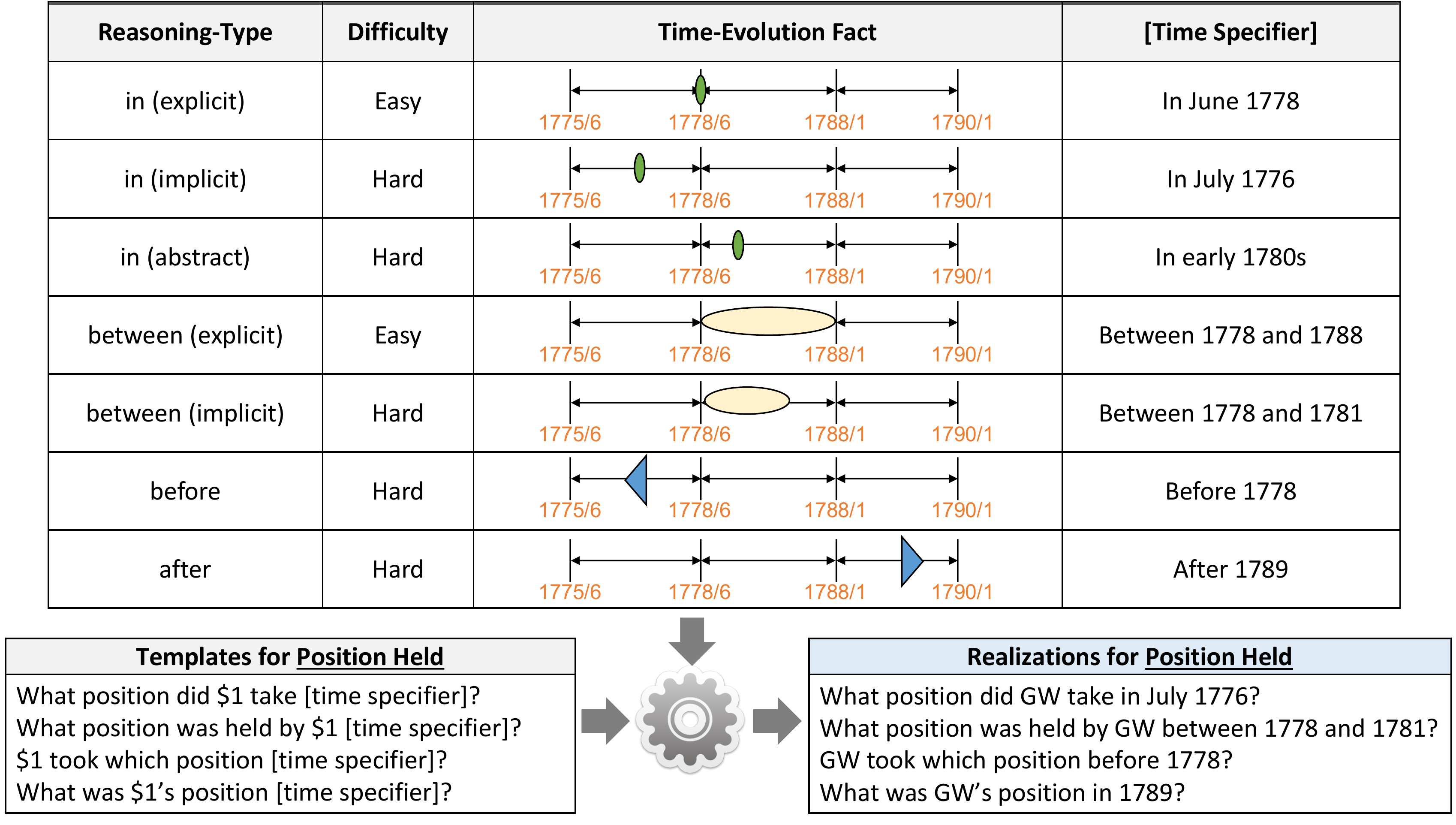}
    \caption{QA Synthesizing Step: taking templates and annotated time-evolving fact as input to generate realistic question-answer pairs. } 
    \label{fig:synthesize}
\end{figure}
\subsection{Synthesizing Question-Answer Pairs}
Once we obtain the human-corrected time-evolving facts, the following step is to generate question-answer pairs from these facts.
\paragraph{Main Dataset}
This is the main dataset we will use throughout our paper, which relies mainly on template generation. The synthesizing procedure is described in~\autoref{fig:synthesize}. For each given relation, we manually write 2-5 different templates. 

We propose several common reasoning types (`in', `between', `before', `after') to fill in the placeholder. For example, if we want to generate `in (implicit)' type, we will randomly sample a time `July 1776' within the segment `1775/6 - 1778/6' and create `in July 1776' to fill in the placeholder. 

As shown in~\autoref{fig:synthesize}, we classify these types into `easy' or `hard' categories depending on whether the [time specifier] exactly matches the boundaries in the time-evolving axis. Since these boundaries are more likely to be mentioned in the passage explicitly, the questions with such [time specifier] on the boundary are easier to be answer based on surface form rather than temporal reasoning. In contrast, [time specifier] falling in the middle of the time span are more likely to necessitate reasoning over the implicit time information. 

To better diagnose the model's capability to perform different levels of temporal reasoning, we generate two versions of TimeQA dataset. In the easy version, we only sample from `easy' reasoning types. In the hard version, we only sample from `hard' reasoning types. In total, we generate 20K questions (average 4 questions per time-evolving fact), for both versions. The easy questions tend to have more explicit mentions in the document, while the hard questions containing more implicit mentions, which is more challenging for the QA models. In the following experiments, we will demonstrate the performance difference between these two versions. The comprehensive statistics of both the easy and hard versions of TimeQA are shown in~\autoref{tab:dataset}. The license and privacy information for the dataset are shown in the Appendix for reference. 

\begin{table}[!t]
\small
\begin{tabular}{llcccccc}
\toprule
Split                  & Mode & \#Questions & \#Entities & \#Relations & \#Answerable & \#Unanswerable & \#Doc-Token \\
\midrule
\multirow{2}{*}{Train} & Easy & 14308       & 3500       & 70          & 12532        & 1776           & 1816         \\
                       & Hard & 14681       & 3500       & 70          & 12532        & 2149           & 1812         \\
\midrule
\multirow{2}{*}{Dev}   & Easy & 3021        & 748        & 52          & 2674         & 347            & 1871         \\
                       & Hard & 3087        & 748        & 52          & 2674         & 413            & 1871         \\
\midrule
\multirow{2}{*}{Test}  & Easy & 2997        & 749        & 50          & 2613         & 384            & 1864         \\
                       & Hard & 3078        & 749        & 50          & 2613         & 465            & 1865         \\
\bottomrule
\end{tabular}
\vspace{2ex}
\caption{The dataset statistics for different splits and modes, \#Doc-Token means the average number of tokens within the document.}
\label{tab:dataset}
\vspace{-2ex}
\end{table}

\begin{figure}[!h]
    \centering
	\includegraphics[width=1.0\linewidth]{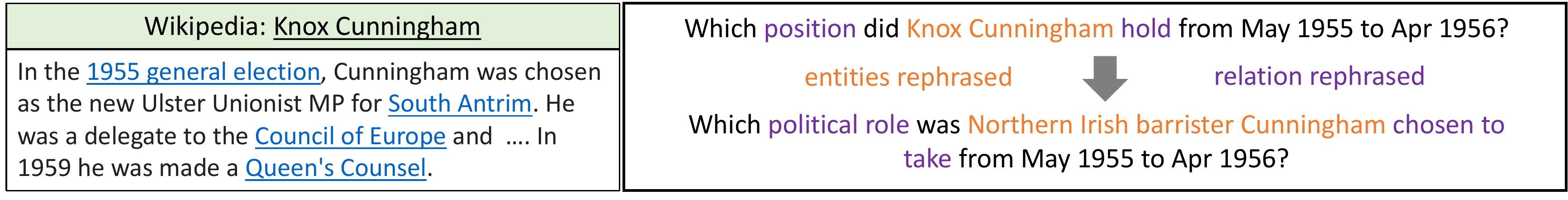}
	\vspace{-2ex}
    \caption{Crowd-Worker Paraphrasing the template questions. } 
    \label{fig:paraphrasing}
\end{figure}

\paragraph{Human-Paraphrased Complement}
To further complement the template-generated questions, we sample a relation-balanced subset of train/test questions for human paraphrasing. In this paraphrasing process, the crowd-workers are required to rewrite the template questions to make them more natural, unambiguous, and diverse. Specifically, the human-written questions will include diverse mentions over the existing relations and entities as shown in~\autoref{fig:paraphrasing}. 
We finally obtain an extra 1171 easy/hard questions regarding 320 time-evolving facts as our training data and 989 easy/hard questions regarding 257 time-evolving facts as our test data. The relations for this human-annotated subset are well balanced to avoid excessive over-fitting.

\section{Models}
Here we formally define the problem setup. The model is given the document $D = d_1, \cdots, d_N$ and question $Q=q_1, \cdots, q_M$, where $q_i$ and $d_i$ refers to $i$-th token in document and question with a length of N and M. The model needs to predict an answer string $\hat{A}$. To cope with the existing challenges, especially the long-term dependency, we propose to use two models BigBird~\citep{zaheer2020big} and FiD~\citep{izacard2020leveraging}, which are known to achieve state-of-the-art performance on the Natural Question~\citep{kwiatkowski2019natural} and TriviaQA~\citep{joshi2017triviaqa}. We briefly describe their design as follows:

\begin{figure}[!htb]
    \centering
	\includegraphics[width=0.9\linewidth]{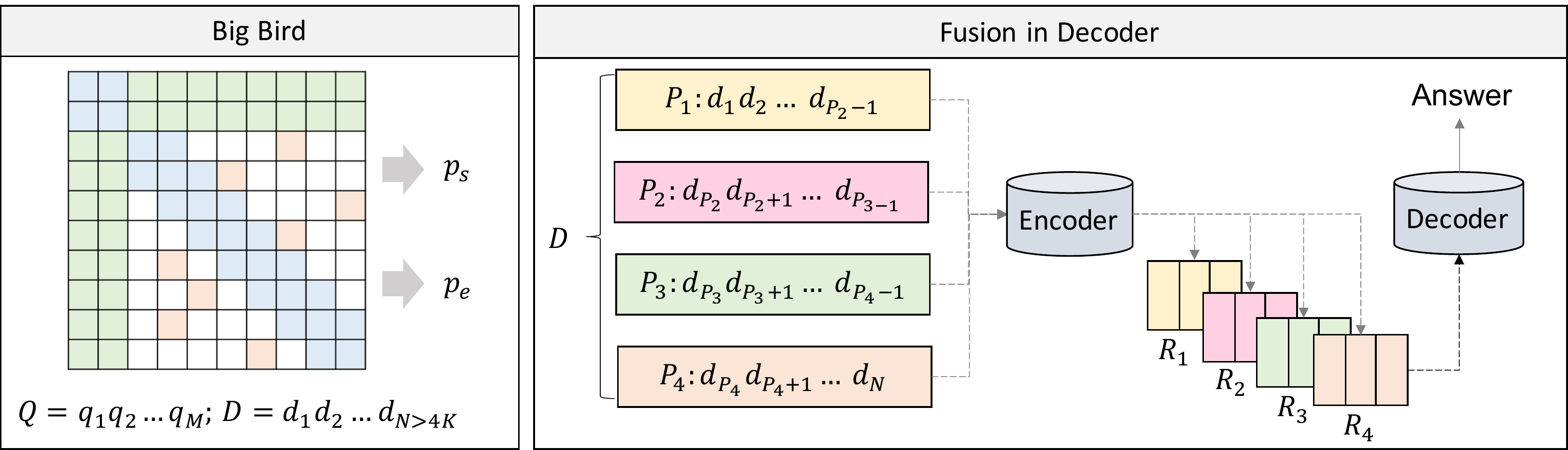}
    \caption{The extractive and generative architectures. (Left: BigBird; Right: FiD)} 
    \label{fig:model}
\end{figure}
\subsection{BigBird Extractive Model}
This model aims to extract the start and end positions from the given sequence. The input sequence $X=(q_1, \cdots, q_M, [SEP], d_1, \cdots, d_N)$ is a concatenated sequence of question $Q$ and document $D$. Due to the length of the document, the input sequence can easily exceed 4K tokens. Therefore, BigBird~\citep{zaheer2020big} uses a more generalized attention mechanism as:
\begin{equation}
    ATTN(X)_i = x_i + \sum_{h=1}^H (\sigma(Q_h(x_i)K_h(X_{N(i)})^T)) \cdot V_h(X_{N(i)})
\end{equation}
where $Q, K, V$ are the query, key and value functions, and $X_{N(i)}$ corresponds to the matrix formed by only stacking ${x_j : j \in N(i)}$ and not all the inputs. BigBird~\citep{zaheer2020big} proposes two attention mechanism: 1) a local sliding window to ensure the nodes $x_i$ attend to its local neighbors $x_{i-w/2:i+w/2}$, 2) a random subset of nodes are also being attended by $x_i$ to capture global information. The above local-global attention mechanism makes $N(i)$ a sparse matrix, which is depicted in left side of~\autoref{fig:model}. Such sparse attention matrix lowers the square computation cost to almost linear. After stacking multiple layers, we obtain the representation at top layer as $R_X \in \mathbb{R}^{(N + M) \times D}$ with $D$ denoting the hidden dimension. We project $R_X$ to $p_{s} \in \mathbb{R}^{N+M}$ and $p_{e}  \in \mathbb{R}^{N+M}$ using the following equation, where $W_s$ and $W_e$ are both $\mathbb{R}^{D \times 1}$ learnable matrices.
\begin{equation}
    p_s = softmax(squeeze(R_X \cdot W_s)); \quad p_e = softmax(squeeze(R_X \cdot W_e))
\end{equation}
During inference time, we select $i, j = argmax_{i,j} (p_s(i) \times p_e(j))$ as the start and end position of the prediction span, and the answer prediction is $\hat{A} = X_{i:j}$.

\subsection{FiD Generative Model}
This model aims to generate answer token by token in an auto-regressive fashion. In order to accommodate the long document $D$, FiD~\citep{izacard2020leveraging} proposes to split it into $L$ relative short paragraphs denoted as $P_1, P_2, \cdots, P_L$, where each of the $P_i$ is shorter than a length limit of $K$ tokens (here $N \approx (K+M)L$, with $K \gg M$). The question $Q$ and $P_i$ are concatenated to build $i$-th input $\hat{P}_i$, where each of $P_i$ is fed to the T5 encoder function $f_{enc}$ to obtain their corresponding representation as $R_i \in \mathbb{R}^{(K+M) \times D}$. The encoded representations are thus concatenated to build $R = [R_1, R_2, \cdots, R_L] \in \mathbb{R}^{N \times D}$. The decoder $f_{dec}$ then attends to the concatenated $R$ to generate the answer $\hat{A}$ token by token using $p(\hat{a}_i|\hat{a}_{1:i-1}, R)$, which is described as follows:
\begin{equation}
    R_i = f_{enc}(\hat{P}_i); \quad p(\hat{a}_i|\hat{a}_{1:i-1}, R) = softmax(f_{dec}(\hat{a}_{1:i-1}, [R_1, \cdots, R_L]) \cdot W_a)
\end{equation}
where $W_a \in \mathbb{R}^{D \times |V|}$ is the learnable matrix to predict next-word probability over vocabulary $|V|$.

Since the decoder sequence here the attention is much shorter, the attention cost is ignorable. Thus, the total computation complexity is lowered to $O((K+M)^2L) \approx O((K+M)N) \ll O(N^2)$. With the almost linear approximation, the generative model can handle input sequences with a length of 4K tokens. We demonstrate the model architecture on the right side of~\autoref{fig:model}.

\section{Experiments}
\subsection{Experimental Setup}
We conduct all the experiments based on HugginFace Transformer~\citep{wolf-etal-2020-transformers}. The BigBird transformer checkpoints fine-tuned on NQ and TriviaQA are downloaded from \url{vasudevgupta/bigbird-roberta-natural-questions} and \url{google/bigbird-base-trivia-itc}. These two models are based on BigBird-base with 12 layers, 12 attention heads, and a hidden dimension of 768. The maximum position embedding is 4096. Both models use a local block size of 64 and 3 random blocks for global attention. The FiD transformer checkpoints fine-tuned on NQ and TriviaQA are downloaded from \url{https://github.com/facebookresearch/FiD}. The FiD-base model also uses 12 layers of encoder/decoder with 12 attention heads. But its maximum position embedding is limited to 512. Thus, Both models are quite comparable in terms of parameter size. 

We fine-tune all the models using AdamW~\citep{loshchilov2018decoupled} with a learning rate of 2e-5. We fine-tune all the models for 3 epochs and evaluate the performance after each epoch on the dev set to select the best-performing model. The models are trained on 4 Titan RTX GPU with 24G memory with a per-GPU-batch-size of 1. For the results in~\autoref{tab:main_result}, Easy/Hard-Mode means we use the corresponding dataset for both training and evaluation.   

\subsection{Evaluation Metrics}
We follow the HotpotQA~\citep{yang2018hotpotqa} and NQ~\citep{kwiatkowski2019natural} to use exact match and precision/recall and F1 score as the evaluation metrics. Assuming we have evaluation examples $\{Q^{(i)}, D^{(i)}, [A^{(i)}]\}$ for $i=1, \cdots, n$, where $[A^{(i)}]$ is a list of annotated answer strings or NULL. Our prediction function outputs $\hat{A}^{(i)}$ as the answer, which is either a string or NULL. The exact match function $EM(\hat{A}^{(i)}, [A^{(i)}]) = max_j \{ eq(A^{(i)}_j, \hat{A}^{(i)})\}$. For F1 score, we first split each answer $A^{(i)}$ into a set of individual words $\mathbb{A}^{(i)}$, and then compute their recall and precision. The formal definition is $F1(\hat{A}^{(i)}, [A^{(i)}]) = max_j \{ F1(\mathbb{A}^{(i)}_j, \hat{\mathbb{A}}^{(i)})\} \}$. 
To place an upper bound on the metrics introduced above, we distribute the annotated test-set questions to crowd workers to let them provide span answers from the given passage. We compare their predictions with the annotated answers and report the approximated human results in~\autoref{tab:main_result} for comparison.
\subsection{Main Results}
We demonstrate our main results as~\autoref{tab:main_result}. In the first block, we use BigBird (pre-trained with MLM) and FiD (initialized from T5 checkpoint) and only fine-tune them on our TimeQA training set without relying on external NQ/TriviaQA data. Since our dataset size is rather limited, the achieved performance is lower than 20\%. 

In the second block, we probe the performance of BigBird and FiD models fine-tuned on large-scale NQ/TriviaQA data. Since the TimeQA questions are linguistically simple and natural, and also coming from the Wikipedia domain, there exists very little distributional shift. However, these fine-tuned models are only achieving 33\% under easy mode and 27\% under the hard mode, much lower than their performance on NQ/TriviaQA (over 60\%). Such a gap reflects concerning incompetence of these models to deal with temporal reasoning in text.

In the third block, we continue to fine-tune these pre-fine-tuned models on our TimeQA training set. This adaptive fine-tuning greatly enhances the models' capability to perform temporal reasoning. The performance can be significantly boosted to 60\% under easy mode and 45\% under hard mode. However, the best-performing model is still far behind the human performance, especially under hard mode (87\%), which indicates large headroom for future studies.

Throughout the experiments, we observe that the model is consistently getting much lower accuracy under hard mode. The significant 15\% performance drop from easy to hard reflects the incompetence of the models to handle robust temporal reasoning. In contrast, humans are suffering only 2\% drop under the hard mode, which indicates that humans are more robust in temporal reasoning. 

\begin{table}[!t]
\centering
\small
\begin{tabular}{l|ll|ll|ll|ll}
\hline
\multirow{3}{*}{Model}                    & \multicolumn{4}{c}{Easy-Mode}                      & \multicolumn{4}{|c}{Hard-Mode}                      \\
\cline{2-9}
                                          & \multicolumn{2}{|c|}{Dev} & \multicolumn{2}{c}{Test} & \multicolumn{2}{|c|}{Dev} & \multicolumn{2}{c}{Test} \\
\cline{2-9}
                                          & EM         & F1         & EM         & F1           & EM         & F1         & EM         & F1          \\
\hline
BigBird (FT on TimeQA)                   &  16.4        & 27.5       & 16.3        & 27.1        & 11.4        & 20.6        & 11.9        & 20.3        \\
FiD (FT on TimeQA)                  & 15.9        & 27.1       & 15.7        & 28.0        & 10.7        & 19.1       & 10.3        & 19.7         \\
\hline
\hline
BigBird (FT on NQ)         &     28.5    &         40.5   &    28.6    &       39.6     &   26.4       &    36.8     &  25.5      & 35.7         \\
BigBird (FT on TriviaQA)      &   \textbf{33.4}    &     \textbf{42.5}       & \textbf{33.7}           &       \textbf{43.0}           &    \textbf{27.7}   &  \textbf{35.9}      &  \textbf{27.7}    &  \textbf{36.2}     \\
FiD (FT on NQ)                &     22.5       &    32.2        &      23.3      &  32.8       &    15.8    &    24.5    &  16.0     &  24.9        \\
FiD (FT on TriviaQA)          &    23.2      &      34.0    &  23.2          &      33.0       &    13.6    &   22.0     &    13.1       & 21.4              \\
\hline
\hline
BigBird (FT on NQ + TimeQA)       &    50.6        &      59.5      &     51.2       &      60.3       &     40.8       &       49.8     &  42.4          &     50.9        \\
BigBird (FT on TriviaQA + TimeQA) &      50.2      &       58.9     &   50.8         &  59.7           &     40.6       &   47.8         &    40.4        &     47.5         \\
FiD  (FT on NQ + TimeQA)       &     \textbf{59.5}	       &   \textbf{66.9}      &     \textbf{60.5}      &    \textbf{67.9}         &    \textbf{45.3}       &   \textbf{54.3}       &   \textbf{46.8}        &   \textbf{54.6}       \\
FiD (FT on TriviaQA + TimeQA)     &     56.4       &  64.9       &    57.5      &         65.1       &         44.4  &    52.5    &   46.2     &    53.7   \\ 
\hline
Human Worker                           &    -           &    -           &   89.0         &      93.3       &    -       &   -     &     87.0   &    91.1    \\ 
\hline
\end{tabular}
\vspace{2ex}
\caption{Main results for different models on our dataset, we report the EM/F1 scores for both dev/test set under easy and hard mode. All results are averaged by 3  runs.}
\label{tab:main_result}
\vspace{-2ex}
\end{table}

\subsection{Human-Paraphrased Results}
We further provide experimental results on human-paraphrased questions in~\autoref{tab:human_result}. We first evaluate the model fine-tuned on NQ to directly answer the human-paraphrased questions, we can observe a 2-4 points drop in EM score. Then we evaluate the model finetuned on NQ+TimeQA (only containing template questions). Surprisingly, the model without being trained on human-paraphrased questions can generalize very well to these human-paraphrases, suffering only 4-5\% EM drop. After applying the 1K human-paraphrased training examples for adaptation, the gap is further decreased to only 2\% EM score. The narrow performance gap between the realistic human-written questions and artificially synthesized questions indicates that our synthesized dataset is indeed an accurate proxy for estimating the model's capability to solve real-world time-sensitive questions.

Since our questions are highly decontextualized, i.e. the question is non-ambiguous, leading to a unique answer in the world. We follow DrQA~\citep{chen2017reading} to perform open-domain QA, where we first use BM25 retriever to retrieve the most relevant passage from whole Wikipedia dump and then run BigBird to extract the answer. The best retrieval accuracy HITS@1 under easy and hard are 28.8\% and 26.8\%. The best end-task QA accuracy is rather low, achieving roughly 14\% under Test-Easy and 11\% under Test-Hard with lot of headroom for future work.
\begin{table}[!t]
\centering
\small
\begin{tabular}{l|ll|ll|ll|ll}
\hline
\multirow{3}{*}{Model}                    & \multicolumn{4}{c}{Close-Domain QA}                      & \multicolumn{4}{|c}{Open-Domain QA }                      \\
\cline{2-9}
                                          & \multicolumn{2}{|c|}{Test-Easy} & \multicolumn{2}{c}{Test-Hard} & \multicolumn{2}{|c|}{Test-Easy} & \multicolumn{2}{c}{Test-Hard} \\
\cline{2-9}
                                          & EM         & F1         & EM         & F1           & EM         & F1         & EM         & F1          \\
\hline
BigBird (FT on NQ)                   &  26.1       &    35.3   &  21.3       & 30.1        &   8.3      &  11.1        &    6.3    &   8.7     \\
BigBird (FT on NQ + TimeQA)   &    45.5     &   51.7      &  36.5       &       43.6   &    13.9     &  15.7        &   10.4     &   12.7     \\
BigBird (FT on NQ + TimeQA \& Human)                   &      47.6   &       53.7 &    38.8     &      45.9   &    14.3     &       17.7   &   11.0     &      13.0  \\
\hline
\end{tabular}
\vspace{2ex}
\caption{Additional results on human-paraphrased split, we report the EM/F1 scores for both close-domain and open-domain settings. All results are averaged by 3  runs.}
\label{tab:human_result}
\vspace{-2ex}
\end{table}

\subsection{Model Analysis}
\begin{figure}[!t]
    \centering
	\includegraphics[width=1.0\linewidth]{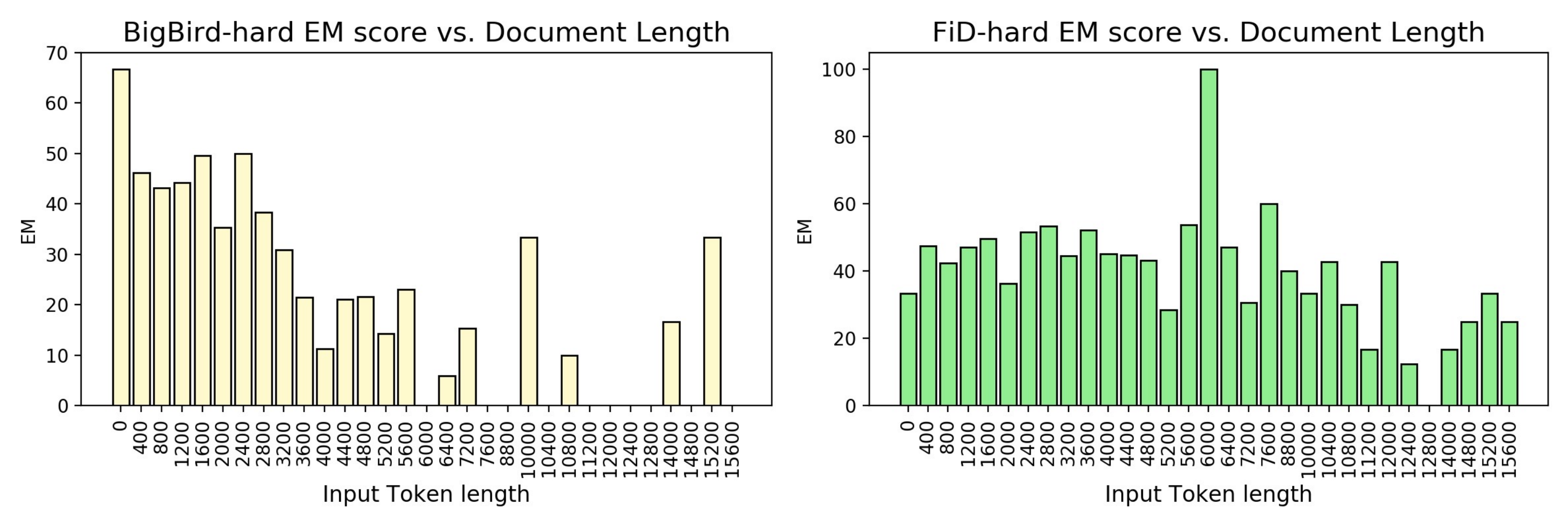}
    \vspace{-5ex}
    \caption{BigBird and FiD's accuracy under length of document input.}
    \label{fig:length}
    \vspace{-3ex}
\end{figure}

\begin{figure}[!t]
    \centering
	\includegraphics[width=1.0\linewidth]{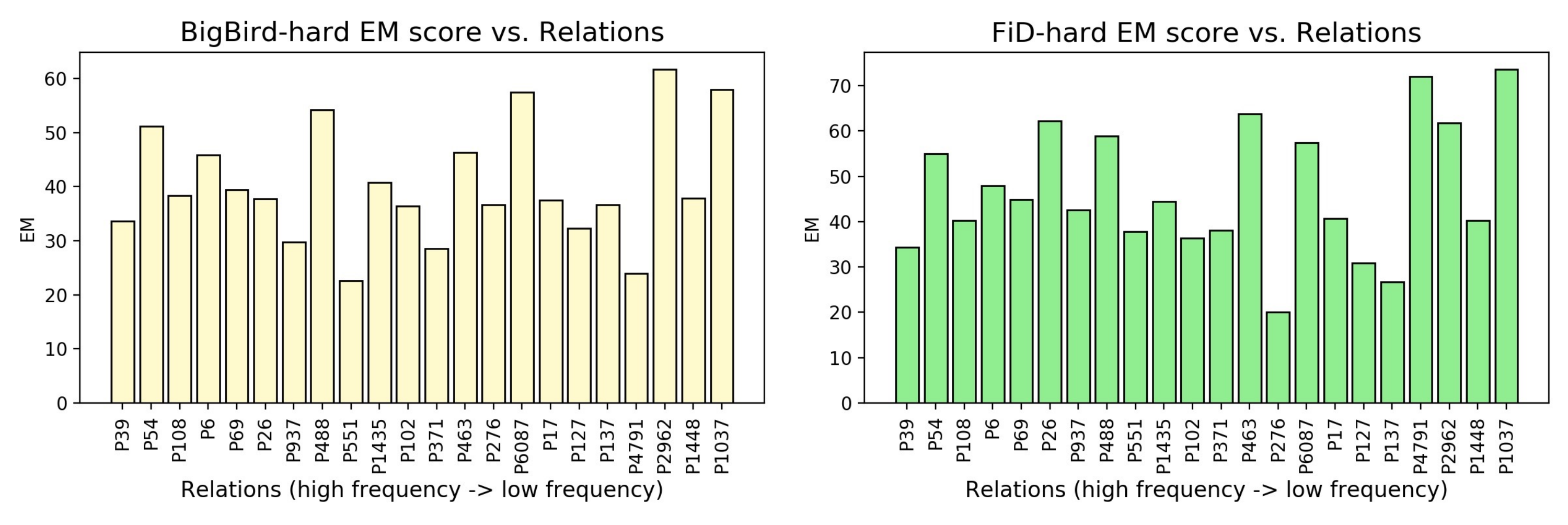}
    \vspace{-5ex}
    \caption{BigBird and FiD's accuracy under different relations.} 
    \label{fig:acc_relations}
    \vspace{-3ex}
\end{figure}

Here we further analyze the model performance from different angles. Specifically, we demonstrate the model's score concerning different relations, and different document length as follows:
\paragraph{Impact of Long-term Dependency}
Here we are interested in understanding the models' capability to handle the long-term dependency in temporal reasoning. We plot the model's accuracy under different document lengths in~\autoref{fig:length}. As can be seen, the BigBird's performance degrades rapidly as the length increases to over 5000 tokens, while the FiD's performance is quite uniformly distributed across different document lengths. This figure demonstrates that the FiD's better performance is partially attributed to its strong capability to deal with long-term dependency in temporal reasoning. 
\paragraph{Impact of Relation Type}
Here we are also interested in understanding the model's performance over different relations and demonstrate our findings in~\autoref{fig:acc_relations}. We found that the model's performance is orthogonal to the frequency of relations. For example, the relation `P1037 (director of)' only has 30 instances, however, its performance is much higher than the relation `P39 (position held)', which has over 500 training instances. It's mainly due to the fact that the time boundary of relation `P1037' is more likely to be explicitly mentioned than `P39 (position held)'.

\paragraph{Consistency Analysis}
We are interested in whether the best-performing models can make consistent predictions under random perturbation of [time specifier]. For example, if a model perceives that a fact $(subject, relation, object)$ persists between $t_s$ and $t_e$, then the model should make consistent prediction for any question with [time specifier] falling within the range of $t_s$ and $t_e$. Therefore, we select the correctly predicted examples and randomly perturb the [time specifier] for 3 times. We observe how many percentages of the model predictions will remain constant/true under these random perturbations. We observe only 66\% of model predictions are agnostic to these perturbations. This finding suggests that the existing QA models are not quite consistent with respect to their predictions.

\subsection{Error Analysis}
To better investigate the errors made by the QA model, we perform a detailed analysis based on the predictions from the best-performing FiD model. Under the easy-dev set, we categorize questions based on whether the [time specifier] is explicitly mentioned in the given passage. We found that two-thirds of easy-questions have [time specifier] explicitly mentioned with the average performance over 64\%, while the rest easy-questions only achieve 49\%. Such comparison indicates that implicit time information is a major challenge to our QA model. We further categorize the errors mainly into sources: 1) temporal reasoning: for example, `... in May 2010, one year after that' refers to `May 2011' based on numeric addition, 2) commonsense reasoning: for example, `... in 2012 London Olympics, in the next Olympic game' refers to `2016 Olympic' based on our commonsense. 3) termination reasoning: the termination time of a fact is commonly unmentioned in a text corpus, it needs to be inferred based on the start time of the next event. For example, `XXX joined A team in 2017, ... in 2019, B team signed a contract with XXX', we know that B team and A team are mutually exclusive, there for the termination time of A team is in 2019.

These three cases are prevalent in our daily text, which poses great challenges for the existing models. To further boost the performance on our dataset, it is vital to consider better algorithms to inject the temporal commonsense knowledge into the models.

\section{Related Work}
\noindent{\bf{Question Answering}}
There have been numerous efforts to tackle the machine reading comprehension problem. Different datasets like DrQA~\citep{chen2017reading}, TriviaQA~\cite{joshi2017triviaqa}, SearchQA~\citep{dunn2017searchqa} and DROP~\citep{dua2019drop} have been proposed. As the SQuAD~\citep{rajpurkar2016squad} questions are relatively simple because they usually require no more than one sentence in the paragraph to answer. The following datasets further challenge the QA model's capability to handle different scenarios like open-domain, long context, multi-hop, discrete operations, etc. However, these QA datasets lack the existence the time-sensitive questions. Thus, we hope our effort could serve as a complement to the existing QA research.\vspace{1ex} \\ 
\noindent{\bf{Temporal Reasoning over Knowledge Base}}
Understanding time evolution is an important research topic. As our world is constantly changing, it's vital to understand the time scope of world knowledge. There have been long-standing efforts to inject temporal quantifiers into the knowledge base (KB) using temporal knowledge extraction techniques~\citep{talukdar2012coupled,chang2012sutime,talukdar2012acquiring,wijaya-etal-2014-ctps}. Adding the temporal information into KB can empower down-stream applications like KBQA~\citep{ahn2006supporting,sanampudi2013question,jia2018tempquestions,jia2018tequila,saxena-etal-2021-question} to handle time-sensitive queries. Such two-step approach might suffer from cascaded errors and lead to compromised accuracy. In contrast, TimeQA aims to directly answer time-sensitive queries based on unstructured text. The new problem is more realistic yet challenging due to the high variance of human expressions over time information. \vspace{1ex} \\
\noindent{\bf{Temporal Reasoning over Text}}
Recently, a contemporary dataset SituatedQA~\citep{zhang2021situatedqa} was also released targeting at answering open-domain time-sensitive QA. There are a few major differences: 1) SituatedQA contains more realistic queries selected from NQ dataset~\citep{kwiatkowski2019natural} while TimeQA contains mostly synthesized queries (except human-paraphrased subset). 2) TimeQA contains 20K queries, while SitutatedQA only contains 4K temporal queries. 3) The hard version of TimeQA requires reasoning over implicit temporal mentions in passage, which is not emphasized in SituatedQA. Another related dataset was introduced by~\cite{dhingra2021time} to diagnose whether the existing LMs are able to sensitive to changes in temporal knowledge. Their questions are mostly cloze-based probing questions under closed-book setting, while both TimeQA and SituatedQA are evaluated under open-book setting. \vspace{1ex} \\
\noindent{\bf{Temporal Reasoning over Events}}
Another popular domain for temporal reasoning is event-centric tasks, which aims at understanding the textual description of real-world events~\citep{chen2021event}. There has been studies on extracting time boundaries for events~\citep{ning-etal-2018-cogcomptime,wen2021event,zhang2021extracting}, reading comprehension over events~\citep{ning2020torque}. These event-centric NLP tasks focus on understanding the temporal relationship (e.g. `before', `after', `include', etc) between multiple events (e.g. `snow melting', `land slide', etc), which have inherent logical relation like `causation', `negation', `inclusion', etc. Compared to TORQUE~\citep{ning2020torque}, there are two major differences: 1) TimeQA requires numerical reasoning over time information while TORQUE does not require it, 2) TimeQA requires modeling long-term temporal dependency in the text, while TORQUE's passages are mostly short text.

\section{Conclusion}
Though time-sensitive facts are pervasive in our daily text corpus, there has been little prior work exploring this direction. In this paper, we build the first dataset to investigate whether existing models can understand time-sensitive facts. Our experiments show that the SoTA models are still lagged behind humans in temporal reasoning. In order to empower the future NLP models to understand temporal information, different temporal-aware models need to be proposed. Finally, this paper opens up new research directions for better modeling temporal information in text representations.

\bibliographystyle{plainnat}
\bibliography{neurips_data_2021}

\begin{thebibliography}{32}
\providecommand{\natexlab}[1]{#1}
\providecommand{\url}[1]{\texttt{#1}}
\expandafter\ifx\csname urlstyle\endcsname\relax
  \providecommand{\doi}[1]{doi: #1}\else
  \providecommand{\doi}{doi: \begingroup \urlstyle{rm}\Url}\fi

\bibitem[Ahn et~al.(2006)Ahn, Schockaert, De~Cock, and
  Kerre]{ahn2006supporting}
David Ahn, Steven Schockaert, Martine De~Cock, and Etienne Kerre.
\newblock Supporting temporal question answering: Strategies for offline data
  collection.
\newblock In \emph{Proceedings of the Fifth International Workshop on Inference
  in Computational Semantics (ICoS-5)}, 2006.

\bibitem[Allen(1983)]{allen1983maintaining}
James~F Allen.
\newblock Maintaining knowledge about temporal intervals.
\newblock \emph{Communications of the ACM}, 26\penalty0 (11):\penalty0
  832--843, 1983.

\bibitem[Bao et~al.(2016)Bao, Duan, Yan, Zhou, and Zhao]{bao2016constraint}
Junwei Bao, Nan Duan, Zhao Yan, Ming Zhou, and Tiejun Zhao.
\newblock Constraint-based question answering with knowledge graph.
\newblock In \emph{Proceedings of COLING 2016, the 26th International
  Conference on Computational Linguistics: Technical Papers}, pages 2503--2514,
  2016.

\bibitem[Berant et~al.(2013)Berant, Chou, Frostig, and
  Liang]{berant2013semantic}
Jonathan Berant, Andrew Chou, Roy Frostig, and Percy Liang.
\newblock Semantic parsing on freebase from question-answer pairs.
\newblock In \emph{Proceedings of the 2013 conference on empirical methods in
  natural language processing}, pages 1533--1544, 2013.

\bibitem[Chang and Manning(2012)]{chang2012sutime}
Angel~X Chang and Christopher~D Manning.
\newblock Sutime: A library for recognizing and normalizing time expressions.
\newblock In \emph{Lrec}, volume 3735, page 3740, 2012.

\bibitem[Chen et~al.(2017)Chen, Fisch, Weston, and Bordes]{chen2017reading}
Danqi Chen, Adam Fisch, Jason Weston, and Antoine Bordes.
\newblock Reading wikipedia to answer open-domain questions.
\newblock In \emph{Proceedings of the 55th Annual Meeting of the Association
  for Computational Linguistics (Volume 1: Long Papers)}, pages 1870--1879,
  2017.

\bibitem[Chen et~al.(2021)Chen, Zhang, Ning, Li, Ji, McKeown, and
  Roth]{chen2021event}
Muhao Chen, Hongming Zhang, Qiang Ning, Manling Li, Heng Ji, Kathleen McKeown,
  and Dan Roth.
\newblock Event-centric natural language understanding.
\newblock In \emph{Proceedings of the 59th Annual Meeting of the Association
  for Computational Linguistics}, 2021.

\bibitem[Dhingra et~al.(2021)Dhingra, Cole, Eisenschlos, Gillick, Eisenstein,
  and Cohen]{dhingra2021time}
Bhuwan Dhingra, Jeremy~R Cole, Julian~Martin Eisenschlos, Daniel Gillick, Jacob
  Eisenstein, and William~W Cohen.
\newblock Time-aware language models as temporal knowledge bases.
\newblock \emph{arXiv preprint arXiv:2106.15110}, 2021.

\bibitem[Dua et~al.(2019)Dua, Wang, Dasigi, Stanovsky, Singh, and
  Gardner]{dua2019drop}
Dheeru Dua, Yizhong Wang, Pradeep Dasigi, Gabriel Stanovsky, Sameer Singh, and
  Matt Gardner.
\newblock Drop: A reading comprehension benchmark requiring discrete reasoning
  over paragraphs.
\newblock In \emph{Proceedings of the 2019 Conference of the North American
  Chapter of the Association for Computational Linguistics: Human Language
  Technologies, Volume 1 (Long and Short Papers)}, pages 2368--2378, 2019.

\bibitem[Dunn et~al.(2017)Dunn, Sagun, Higgins, Guney, Cirik, and
  Cho]{dunn2017searchqa}
Matthew Dunn, Levent Sagun, Mike Higgins, V~Ugur Guney, Volkan Cirik, and
  Kyunghyun Cho.
\newblock Searchqa: A new q\&a dataset augmented with context from a search
  engine.
\newblock \emph{arXiv preprint arXiv:1704.05179}, 2017.

\bibitem[Izacard and Grave(2020)]{izacard2020leveraging}
Gautier Izacard and Edouard Grave.
\newblock Leveraging passage retrieval with generative models for open domain
  question answering.
\newblock \emph{arXiv preprint arXiv:2007.01282}, 2020.

\bibitem[Jia et~al.(2018{\natexlab{a}})Jia, Abujabal, Saha~Roy, Str{\"o}tgen,
  and Weikum]{jia2018tempquestions}
Zhen Jia, Abdalghani Abujabal, Rishiraj Saha~Roy, Jannik Str{\"o}tgen, and
  Gerhard Weikum.
\newblock Tempquestions: A benchmark for temporal question answering.
\newblock In \emph{Companion Proceedings of the The Web Conference 2018}, pages
  1057--1062, 2018{\natexlab{a}}.

\bibitem[Jia et~al.(2018{\natexlab{b}})Jia, Abujabal, Saha~Roy, Str{\"o}tgen,
  and Weikum]{jia2018tequila}
Zhen Jia, Abdalghani Abujabal, Rishiraj Saha~Roy, Jannik Str{\"o}tgen, and
  Gerhard Weikum.
\newblock Tequila: Temporal question answering over knowledge bases.
\newblock In \emph{Proceedings of the 27th ACM International Conference on
  Information and Knowledge Management}, pages 1807--1810, 2018{\natexlab{b}}.

\bibitem[Joshi et~al.(2017)Joshi, Choi, Weld, and
  Zettlemoyer]{joshi2017triviaqa}
Mandar Joshi, Eunsol Choi, Daniel~S Weld, and Luke Zettlemoyer.
\newblock Triviaqa: A large scale distantly supervised challenge dataset for
  reading comprehension.
\newblock In \emph{Proceedings of the 55th Annual Meeting of the Association
  for Computational Linguistics (Volume 1: Long Papers)}, pages 1601--1611,
  2017.

\bibitem[Kwiatkowski et~al.(2019)Kwiatkowski, Palomaki, Redfield, Collins,
  Parikh, Alberti, Epstein, Polosukhin, Devlin, Lee,
  et~al.]{kwiatkowski2019natural}
Tom Kwiatkowski, Jennimaria Palomaki, Olivia Redfield, Michael Collins, Ankur
  Parikh, Chris Alberti, Danielle Epstein, Illia Polosukhin, Jacob Devlin,
  Kenton Lee, et~al.
\newblock Natural questions: a benchmark for question answering research.
\newblock \emph{Transactions of the Association for Computational Linguistics},
  7:\penalty0 453--466, 2019.

\bibitem[Loshchilov and Hutter(2018)]{loshchilov2018decoupled}
Ilya Loshchilov and Frank Hutter.
\newblock Decoupled weight decay regularization.
\newblock In \emph{International Conference on Learning Representations}, 2018.

\bibitem[Ning et~al.(2018)Ning, Zhou, Feng, Peng, and
  Roth]{ning-etal-2018-cogcomptime}
Qiang Ning, Ben Zhou, Zhili Feng, Haoruo Peng, and Dan Roth.
\newblock {C}og{C}omp{T}ime: A tool for understanding time in natural language.
\newblock In \emph{Proceedings of the 2018 Conference on Empirical Methods in
  Natural Language Processing: System Demonstrations}, pages 72--77, Brussels,
  Belgium, November 2018. Association for Computational Linguistics.
\newblock \doi{10.18653/v1/D18-2013}.
\newblock URL \url{https://aclanthology.org/D18-2013}.

\bibitem[Ning et~al.(2020)Ning, Wu, Han, Peng, Gardner, and
  Roth]{ning2020torque}
Qiang Ning, Hao Wu, Rujun Han, Nanyun Peng, Matt Gardner, and Dan Roth.
\newblock Torque: A reading comprehension dataset of temporal ordering
  questions.
\newblock In \emph{Proceedings of the 2020 Conference on Empirical Methods in
  Natural Language Processing (EMNLP)}, pages 1158--1172, 2020.

\bibitem[Rajpurkar et~al.(2016)Rajpurkar, Zhang, Lopyrev, and
  Liang]{rajpurkar2016squad}
Pranav Rajpurkar, Jian Zhang, Konstantin Lopyrev, and Percy Liang.
\newblock Squad: 100,000+ questions for machine comprehension of text.
\newblock In \emph{Proceedings of the 2016 Conference on Empirical Methods in
  Natural Language Processing}, pages 2383--2392, 2016.

\bibitem[Rajpurkar et~al.(2018)Rajpurkar, Jia, and Liang]{rajpurkar2018know}
Pranav Rajpurkar, Robin Jia, and Percy Liang.
\newblock Know what you don’t know: Unanswerable questions for squad.
\newblock In \emph{Proceedings of the 56th Annual Meeting of the Association
  for Computational Linguistics (Volume 2: Short Papers)}, pages 784--789,
  2018.

\bibitem[Sanampudi and Guda(2013)]{sanampudi2013question}
Suresh~Kumar Sanampudi and Vanitha Guda.
\newblock A question answering system supporting temporal queries.
\newblock In \emph{International Conference on Advances in Computing,
  Communication and Control}, pages 207--214. Springer, 2013.

\bibitem[Saxena et~al.(2021)Saxena, Chakrabarti, and
  Talukdar]{saxena-etal-2021-question}
Apoorv Saxena, Soumen Chakrabarti, and Partha Talukdar.
\newblock Question answering over temporal knowledge graphs.
\newblock In \emph{Proceedings of the 59th Annual Meeting of the Association
  for Computational Linguistics and the 11th International Joint Conference on
  Natural Language Processing (Volume 1: Long Papers)}, pages 6663--6676,
  Online, August 2021. Association for Computational Linguistics.
\newblock \doi{10.18653/v1/2021.acl-long.520}.
\newblock URL \url{https://aclanthology.org/2021.acl-long.520}.

\bibitem[Talukdar et~al.(2012{\natexlab{a}})Talukdar, Wijaya, and
  Mitchell]{talukdar2012acquiring}
Partha~Pratim Talukdar, Derry Wijaya, and Tom Mitchell.
\newblock Acquiring temporal constraints between relations.
\newblock In \emph{Proceedings of the 21st ACM international conference on
  Information and knowledge management}, pages 992--1001, 2012{\natexlab{a}}.

\bibitem[Talukdar et~al.(2012{\natexlab{b}})Talukdar, Wijaya, and
  Mitchell]{talukdar2012coupled}
Partha~Pratim Talukdar, Derry Wijaya, and Tom Mitchell.
\newblock Coupled temporal scoping of relational facts.
\newblock In \emph{Proceedings of the fifth ACM international conference on Web
  search and data mining}, pages 73--82, 2012{\natexlab{b}}.

\bibitem[Vrande{\v{c}}i{\'c} and Kr{\"o}tzsch(2014)]{vrandevcic2014wikidata}
Denny Vrande{\v{c}}i{\'c} and Markus Kr{\"o}tzsch.
\newblock Wikidata: a free collaborative knowledgebase.
\newblock \emph{Communications of the ACM}, 57\penalty0 (10):\penalty0 78--85,
  2014.

\bibitem[Wen et~al.(2021)Wen, Qu, Ji, Ning, Han, Sil, Tong, and
  Roth]{wen2021event}
Haoyang Wen, Yanru Qu, Heng Ji, Qiang Ning, Jiawei Han, Avirup Sil, Hanghang
  Tong, and Dan Roth.
\newblock Event time extraction and propagation via graph attention networks.
\newblock In \emph{Proceedings of the 2021 Conference of the North American
  Chapter of the Association for Computational Linguistics: Human Language
  Technologies}, pages 62--73, 2021.

\bibitem[Wijaya et~al.(2014)Wijaya, Nakashole, and
  Mitchell]{wijaya-etal-2014-ctps}
Derry~Tanti Wijaya, Ndapandula Nakashole, and Tom~M. Mitchell.
\newblock {CTP}s: Contextual temporal profiles for time scoping facts using
  state change detection.
\newblock In \emph{Proceedings of the 2014 Conference on Empirical Methods in
  Natural Language Processing ({EMNLP})}, pages 1930--1936, Doha, Qatar,
  October 2014. Association for Computational Linguistics.
\newblock \doi{10.3115/v1/D14-1207}.
\newblock URL \url{https://aclanthology.org/D14-1207}.

\bibitem[Wolf et~al.(2020)Wolf, Debut, Sanh, Chaumond, Delangue, Moi, Cistac,
  Rault, Louf, Funtowicz, Davison, Shleifer, von Platen, Ma, Jernite, Plu, Xu,
  Scao, Gugger, Drame, Lhoest, and Rush]{wolf-etal-2020-transformers}
Thomas Wolf, Lysandre Debut, Victor Sanh, Julien Chaumond, Clement Delangue,
  Anthony Moi, Pierric Cistac, Tim Rault, Rémi Louf, Morgan Funtowicz, Joe
  Davison, Sam Shleifer, Patrick von Platen, Clara Ma, Yacine Jernite, Julien
  Plu, Canwen Xu, Teven~Le Scao, Sylvain Gugger, Mariama Drame, Quentin Lhoest,
  and Alexander~M. Rush.
\newblock Transformers: State-of-the-art natural language processing.
\newblock In \emph{Proceedings of the 2020 Conference on Empirical Methods in
  Natural Language Processing: System Demonstrations}, pages 38--45, Online,
  October 2020. Association for Computational Linguistics.
\newblock URL \url{https://www.aclweb.org/anthology/2020.emnlp-demos.6}.

\bibitem[Yang et~al.(2018)Yang, Qi, Zhang, Bengio, Cohen, Salakhutdinov, and
  Manning]{yang2018hotpotqa}
Zhilin Yang, Peng Qi, Saizheng Zhang, Yoshua Bengio, William Cohen, Ruslan
  Salakhutdinov, and Christopher~D Manning.
\newblock Hotpotqa: A dataset for diverse, explainable multi-hop question
  answering.
\newblock In \emph{Proceedings of the 2018 Conference on Empirical Methods in
  Natural Language Processing}, pages 2369--2380, 2018.

\bibitem[Zaheer et~al.(2020)Zaheer, Guruganesh, Dubey, Ainslie, Alberti,
  Ontanon, Pham, Ravula, Wang, Yang, et~al.]{zaheer2020big}
Manzil Zaheer, Guru Guruganesh, Kumar~Avinava Dubey, Joshua Ainslie, Chris
  Alberti, Santiago Ontanon, Philip Pham, Anirudh Ravula, Qifan Wang, Li~Yang,
  et~al.
\newblock Big bird: Transformers for longer sequences.
\newblock In \emph{NeurIPS}, 2020.

\bibitem[Zhang and Choi(2021)]{zhang2021situatedqa}
Michael~JQ Zhang and Eunsol Choi.
\newblock Situatedqa: Incorporating extra-linguistic contexts into qa.
\newblock \emph{arXiv preprint arXiv:2109.06157}, 2021.

\bibitem[Zhang et~al.(2021)Zhang, Huang, and Ning]{zhang2021extracting}
Shuaicheng Zhang, Lifu Huang, and Qiang Ning.
\newblock Extracting temporal event relation with syntactic-guided temporal
  graph transformer.
\newblock \emph{arXiv preprint arXiv:2104.09570}, 2021.

\end{thebibliography}

\clearpage
\appendix

\section{Appendix}
\subsection{Dataset documentation and intended uses}
We follow datasheets for datasets guideline to document the followings.

\subsubsection{Motivation}
\begin{itemize}
    \item For what purpose was the dataset created? Was there a speciﬁc task in mind? Was there a speciﬁc gap that needed to be filled? \\ TimeQA is created to test current models' capability to perform diverse temporal reasoning under unstructured text corpus, which can help future NLP models to better capture the time dimension.
    \item Who created the dataset (e.g., which team, research group) and on behalf of which entity (e.g., company, institution, organization)?\\
    UCSB NLP team, mostly Wenhu Chen
\end{itemize}
\subsubsection{Composition}
\begin{itemize}
    \item What do the instances that comprise the dataset represent (e.g., documents, photos, people, countries)? Are there multiple types of instances (e.g., movies, users, and ratings; people and interactions between them; nodes and edges)? \\
    TimeQA only contains documents (text) in the dataset.
    \item How many instances are there in total (of each type, if appropriate)?\\
    There are roughly 20K question-answer pairs.
    \item Does the dataset contain all possible instances or is it a sample (not necessarily random) of instances from a larger set? If the dataset is a sample, then what is the larger set? Is the sample representative of the larger set (e.g., geographic coverage)? If so, please describe how this representativeness was validated/veriﬁed. If it is not representative of the larger set, please describe why not (e.g., to cover a more diverse range of instances, because instances were withheld or unavailable). \\
    It's sampled from large Wikipedia passages, it's representative of all the possible temporal-sensitive information.
    \item Are relationships between individual instances made explicitly (e.g., users’ movie ratings, social network links)? If so, please describe how these relationships are made explicit. \\
    N/A.
    \item Are there recommended data splits (e.g., training, development/validation, testing)? If so, please provide a description of these splits, explaining the rationale behind them.\\
    Yes, we split training, development, and testing set. We split randomly within each data source.
    \item Are there any errors, sources of noise, or redundancies in the dataset? If so, please provide a description.\\
    There could have some potential noise of question or answer annotation.
    \item Is the dataset self-contained, or does it link to or otherwise rely on external resources (e.g., websites, tweets, other datasets)? If it links to or relies on external resources, a) are there guarantees that they will exist, and remain constant, over time; b) are there ofﬁcial archival versions of the complete dataset (i.e., including the external resources as they existed at the time the dataset was created); c) are there any restrictions] (e.g., licenses, fees) associated with any of the external resources that might apply to a future user? Please provide descriptions of all external resources and any restrictions associated with them, as well as links or other access points, as appropriate.\\
    TimeQA is self-contained.
    \item Does the dataset contain data that might be considered conﬁdential (e.g., data that is protected by legal privilege or by doctorpatient conﬁdentiality, data that includes the content of individuals’ non-public communications)? If so, please provide a description.\\
    No, all the samples in TimeQA is public available.
    \item Does the dataset contain data that, if viewed directly, might be offensive, insulting, threatening, or might otherwise cause anxiety? If so, please describe why.\\
    No
    \item Does the dataset relate to people? If not, you may skip the remaining questions in this section.\\
    No
\end{itemize}

\subsubsection{Uses}
\begin{itemize}
    \item Has the dataset been used for any tasks already? If so, please provide a description?\\
    It is proposed to use for QA task.
    \item Is there a repository that links to any or all papers or systems that use the dataset? If so, please provide a link or other access point.\\
    It is a new dataset. We run existing state-of-the-art models and release the code at https: https://github.com/wenhuchen/Time-Sensitive-QA
    \item What (other) tasks could the dataset be used for?\\
    Many other tasks like relation extractions can be also used.
    \item Is there anything about the composition of the dataset or the way it was collected and preprocessed/cleaned/labeled that might impact future uses? For example, is there anything that a future user might need to know to avoid uses that could result in unfair treatment of individuals or groups (e.g., stereotyping, quality of service issues) or other undesirable harms (e.g., ﬁnancial harms, legal risks) If so, please provide a description. Is there anything a future user could do to mitigate these undesirable harms?\\
    N/A
    \item Are there tasks for which the dataset should not be used? If so, please provide a description.
    N/A
\end{itemize}

\subsubsection{Distribution}
\begin{itemize}
    \item Will the dataset be distributed to third parties outside of the entity (e.g., company, institution, organization) on behalf of which the dataset was created? If so, please provide a description.\\
    No.
    \item How will the dataset will be distributed (e.g., tarball on website, API, GitHub)? Does the dataset have a digital object identiﬁer (DOI)?\\
    Release on Github. No DOI
    \item When will the dataset be distributed?\\
    It is released in https://github.com/wenhuchen/Time-Sensitive-QA
    \item Will the dataset be distributed under a copyright or other intellectual property (IP) license, and/or under applicable terms of use (ToU)? If so, please describe this license and/or ToU, and provide a link or other access point to, or otherwise reproduce, any relevant licensing terms or ToU, as well as any fees associated with these restrictions.\\
    BSD 3-Clause "New" or "Revised" License. https://github.com/wenhuchen/Time-Sensitive-QA/blob/main/LICENSE
    \item Have any third parties imposed IP-based or other restrictions on the data associated with the instances? If so, please describe these restrictions, and provide a link or other access point to, or otherwise reproduce, any relevant licensing terms, as well as any fees associated with these restrictions.\\
    No.
    \item Do any export controls or other regulatory restrictions apply to the dataset or to individual instances? If so, please describe these restrictions, and provide a link or other access point to, or otherwise reproduce, any supporting documentation.
    No.
\end{itemize}

\subsubsection{Accessibility}

\begin{itemize}
    \item Links to access the dataset and its metadata: the github repository https://github.com/wenhuchen/Time-Sensitive-QA.
    \item The data is saved in a json format, where an example is shown in the README.md file.
    \item UCSB NLP group will maintain this dataset on the Github account.
    \item BSD 3-Clause "New" or "Revised" License https://github.com/wenhuchen/Time-Sensitive-QA/blob/main/LICENSE
\end{itemize}

\subsection{Evaluation Metrics}
The F1 score is calculated with the following equation to cover the `[unanswerable]' case.
\begin{align*}
    F1(\mathbb{A}^{(i)}_j, \hat{\mathbb{A}}^{(i)}) = 
    \begin{cases}
    1, & if A^{(i)}_j = NULL \And \hat{A}^{(i)} = NULL\\
    0, & if A^{(i)}_j = NULL \And \hat{A}^{(i)} \neq NULL\\
    0, & if A^{(i)}_j \neq NULL \And \hat{A}^{(i)} = NULL\\
    f1(\mathbb{A}^{(i)}_j, \hat{\mathbb{A}}^{(i)}), & if A^{(i)}_j \neq NULL \And \hat{A}^{(i)} \neq NULL\\
    \end{cases}
\end{align*}

\subsection{Annotation Interface}
\begin{figure}[!htb]
    \centering
	\includegraphics[width=0.9\linewidth]{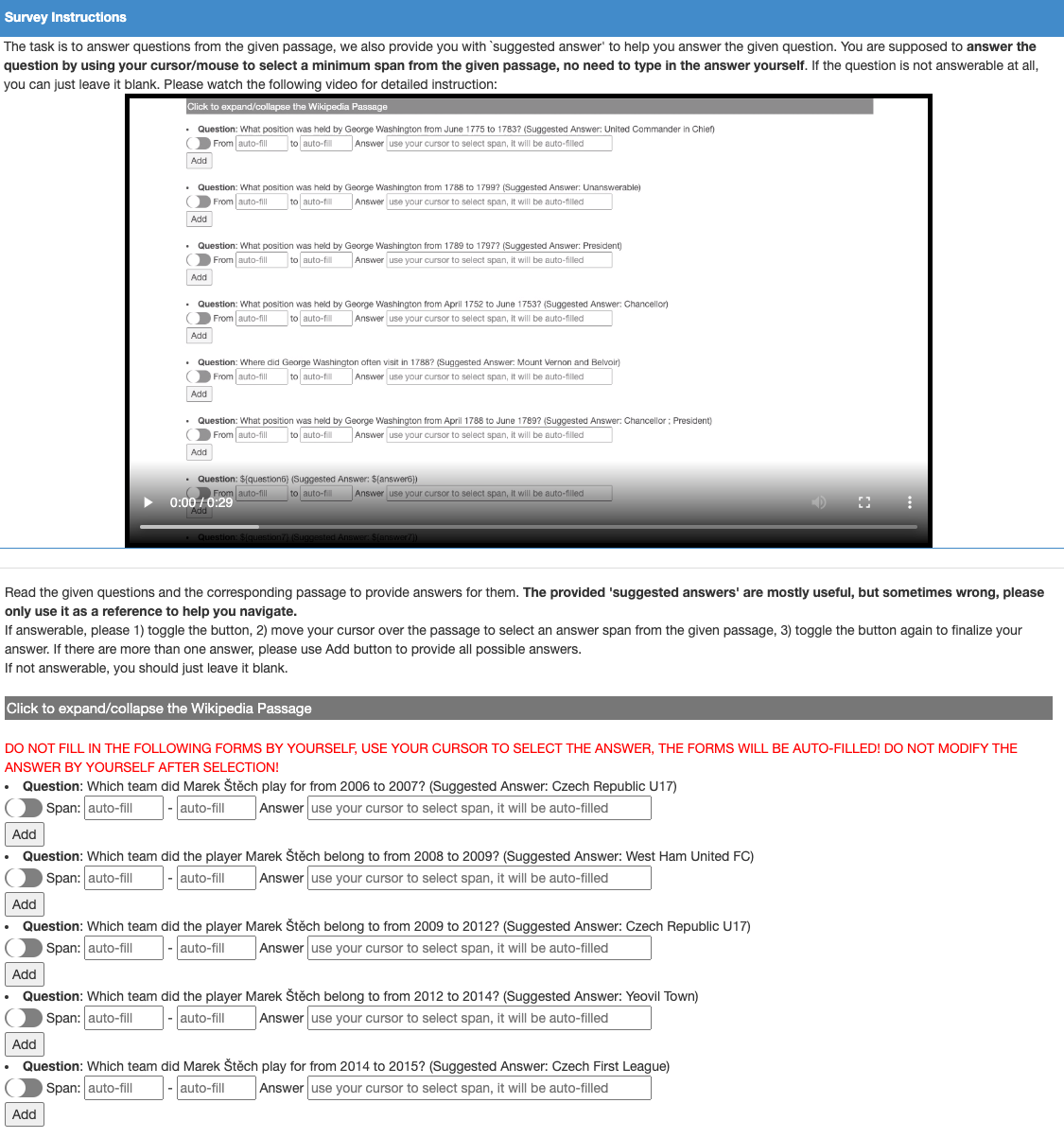}
    \caption{The annotation interface for crowd workers.} 
    \label{fig:interface}
\end{figure}

The annotation interface is demonstrated in~\autoref{fig:interface}, the original HIT job url is \url{https://s3.amazonaws.com/mturk_bulk/hits/467870457/JV_fEwGzVy_PgHqtWqWXLA.html}. 

\subsection{Relation Distribution over TimeQA}
\begin{figure}[!htb]
    \centering
	\includegraphics[width=0.9\linewidth]{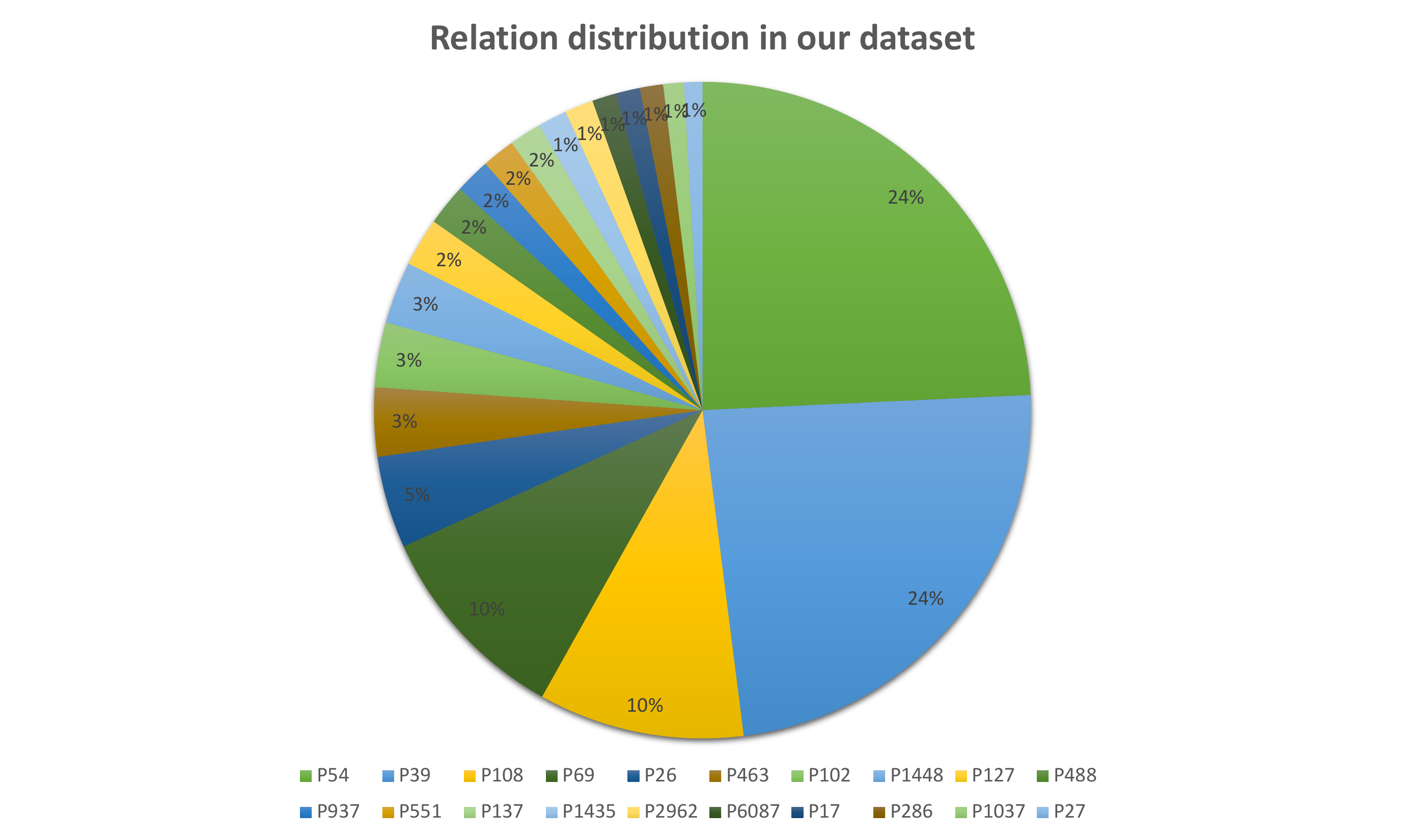}
    \caption{The relation distribution over the annotated facts.} 
    \label{fig:relation}
\end{figure}

The annotated time-evolving facts include more than 70 relations, which follows a long-tail distribution. The major relations are shown in~\autoref{fig:relation}, where the dominant relations are `P54' (play for), `P39' (position held), `P108' (employer of), `P69' (educated at). 

\subsection{Answerable vs. Unanswerable}
\begin{figure}[!htb]
    \centering
	\includegraphics[width=0.9\linewidth]{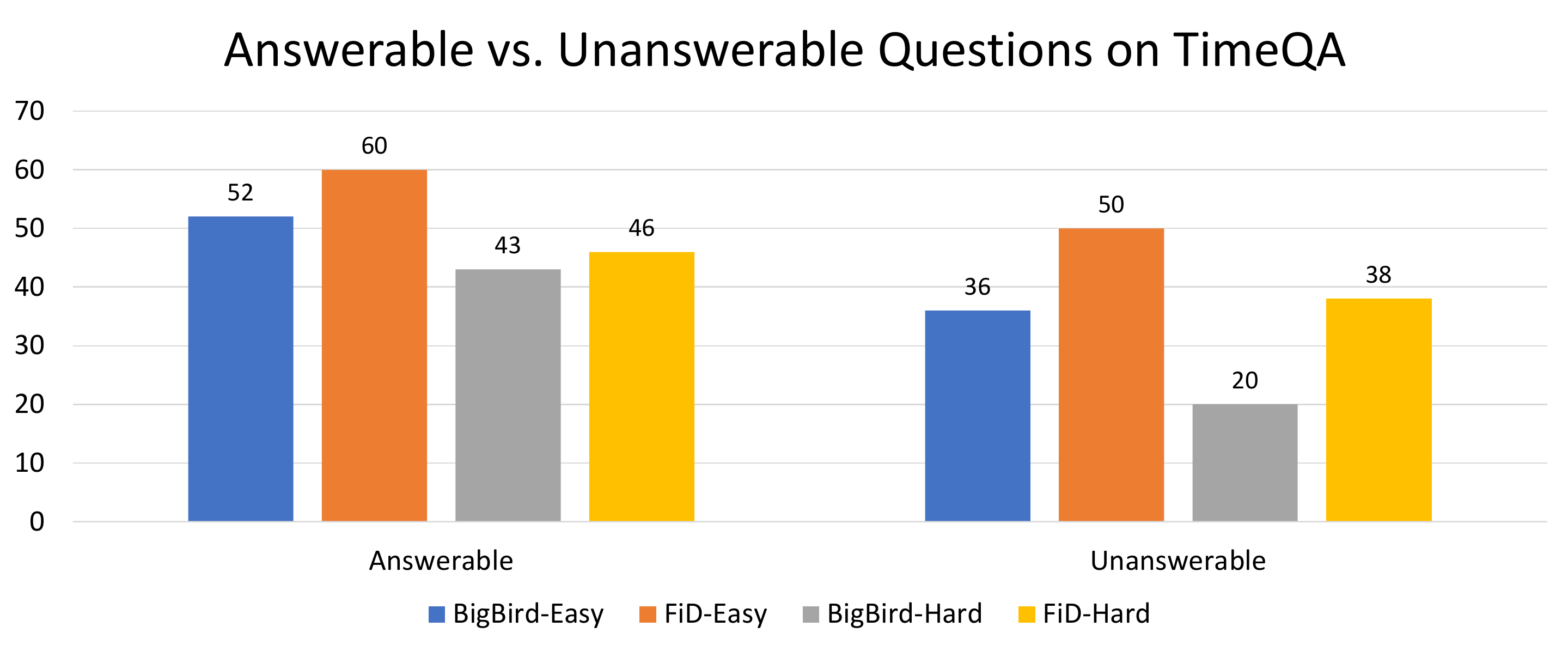}
    \caption{The answerable and unanswerable question performance for BigBird and FiD.} 
    \label{fig:answerable}
\end{figure}

We also provide break-down analysis of model performance for answerable and unanswerable questions in~\autoref{fig:answerable}. As can be seen, the FiD is more aware of the answerability on TimeQA. On the easy and hard mode, FiD's accuracy on unanswerable is above BigBird's by 14\% and 18\%, while the gap in answerable questions are less significant.

\end{document}